\documentclass[10pt,conference]{IEEEtran}
\usepackage{cite}
\usepackage{amsmath,amssymb,amsfonts}
\usepackage{algorithmic}
\usepackage{graphicx}
\usepackage{textcomp}
\usepackage{xcolor}

\PassOptionsToPackage{hyphens}{url}
\usepackage[pdftex,
            pdfauthor={Maxmilian Schall, Tamara Czinczoll, Gerard de Melo},
            pdftitle={CommitBench: A Benchmark for Commit Message Generation}]{hyperref} %
\usepackage{url}           %
\usepackage{booktabs}       %
\usepackage{amsfonts}       %
\usepackage{nicefrac}       %
\usepackage{microtype}      %
\usepackage{xcolor}         %

\usepackage{inconsolata}
\usepackage{multirow}
\usepackage{enumitem}
\usepackage{booktabs}
\setlist{nosep}
\usepackage{makecell}
\usepackage{amssymb}
\usepackage{pifont}
\usepackage{colortbl}
\usepackage[autostyle]{csquotes}
\newcommand{\xmark}{\ding{55}}%

\begin{document}

\title{CommitBench:\\A Benchmark for Commit Message Generation}

\author{\IEEEauthorblockN{%
Maxmilian Schall}
\IEEEauthorblockA{\textit{Hasso Plattner Institute} \\
\textit{University of Potsdam}\\
Potsdam, Germany \\
Maximilian.Schall@hpi.de}
\and
\IEEEauthorblockN{%
Tamara Czinczoll}
\IEEEauthorblockA{\textit{Hasso Plattner Institute} \\
\textit{University of Potsdam}\\
Potsdam, Germany \\
Tamara.Czinczoll@hpi.de}
\and
\IEEEauthorblockN{%
Gerard de Melo}
\IEEEauthorblockA{\textit{Hasso Plattner Institute} \\
\textit{University of Potsdam}\\
Potsdam, Germany \\
Gerard.deMelo@hpi.de}
}

\maketitle

\begin{abstract}
    Writing commit messages is a tedious daily task for many software developers, and
    often remains neglected. Automating this task has the potential to save time while ensuring that messages are informative. A high-quality dataset and an objective benchmark are vital preconditions for solid research and evaluation towards this goal. We show that existing datasets exhibit various problems, such as the quality of the commit selection, small sample sizes, duplicates, privacy issues,
    and missing licenses for redistribution. This can lead to unusable models and skewed evaluations, where inferior models achieve higher evaluation scores due to biases in the data.
    We compile a new large-scale dataset, CommitBench, adopting best practices for dataset creation. We sample commits from diverse projects with licenses that permit redistribution and apply our filtering and dataset enhancements to improve the quality of generated commit messages. We use CommitBench to compare existing models and show that other approaches are outperformed by a Transformer model pretrained on source code. We hope to accelerate future research by publishing the \href{https://github.com/Maxscha/commitbench}{source code}.
\end{abstract}


\section{Introduction}
\label{cha:introduction}

Developers generally appreciate informative commit messages that verbally describe the changes made to a software code base when a new version (commit) is recorded in a version control system (VCS).
While well-crafted commit messages can be of immense help to understand the history and design decisions of a software project, many developers find them tedious to write and thus they are often neglected in practice \cite{maalej_can_2010,dyer_boa_2013}. Another challenge is that developers have different standards, qualitative expectations, and styles regarding the commit message.

A system that automatically suggests informative and consistent commit messages would have the potential to save valuable time and enhance the quality of over 100 million commit messages a day.\footnote{\url{https://octoverse.github.com/2019/} (visited on January 12, 2023) and \url{https://evansdata.com/press/viewRelease.php?pressID=278} (visited on January 12, 2023)}

In recent years, a number of existing artificial intelligence models, originally developed for natural language processing (NLP) tasks such as machine translation or summarization, have been trained to translate code into text~\cite{gros_code_2020, hu_deep_2018, hu_deep_2020} or vice versa~\cite{iyer_mapping_2018, phan_cotext_2021, lu_codexglue_2021}, but also to translate and summarize version differences into commit messages~\cite{liu_neural-machine-translation-based_2018, etemadi_relevance_2020, liu_generating_2019, xu_commit_2019, nie_coregen_2021, liu_atom_2020}. 
Most research on commit message generation is based on datasets that have become outdated, either due to their small size, their low quality, or their lack of concern for privacy, reproducibility, and license restrictions. These limitations point towards the need for better datasets for the task of commit message generation.

We present \texttt{CommitBench}, a new dataset for training and evaluating AI models. This datasets not only adopts existing best practices regarding the repository selection and filters, but also introduces novel filtering techniques. These contributions aim to enhance the quality of the dataset, while considering factors such as privacy, reproducibility, and licenses. We embed CommitBench into the research landscape by training and evaluating models on a selection of existing commit message generation datasets as well as ours.

The dataset is available on Zenodo\footnote{\url{https://zenodo.org/records/10497442}} and HuggingFace\footnote{\url{https://huggingface.co/datasets/maxscha/commitbench}}\footnote{ \url{https://huggingface.co/datasets/maxscha/commitbench_long}}. The source code for the dataset generation is available on GitHub.\footnote{\url{https://github.com/maxscha/commitbench}}

\section{Problem Statement}
Given two versions of the same source code, the first being the original source code and the second an adapted version of the first with $t$ blocks of code changes, a \emph{diff} is a short text that summarizes these changes. A single diff consists of $t$ blocks of changed code lines along with the original ones as well as extra context, typically three lines before and after the change. 

The task of commit message generation can be phrased as a sequence-to-sequence task:
Given a complete textual diff consisting of a sequence of tokens 
$x_1,\dots,x_n$ 
of length $n\in \mathbb{N}$, 
generate a sequence 
$y_1, \dots, y_m$ 
of $m\in \mathbb{N}$ tokens describing the changes in natural language. The descriptions are expected to be informative, yet also short enough to allow developers to quickly skim long commit histories. The input length in the form of source code changes is usually substantially longer than the resulting commit message, i.e., $n \gg m$.

\section{Related Work}
\label{cha:related_work}

\subsection{Commit Messages}
Commit messages play an important role in software engineering.
One study investigated which factors contribute to a high quality commit message, and which are detrimental, according to developers \cite{tian_what_2022}. The study finds that, while there is currently no universal standard for commit messages, most developers agree that not only should the commit message describe \emph{what} changes are made, but also \emph{why} they were done. The authors analyzed five open source software projects and their human-generated commit messages, concluding that on average 44\% of all commit messages are in need of improvement. They published a machine learning model trained on this data to classify good commit messages. 

The negligence of software developers, who often fail to take the time to write high-quality commit messages, along with advances in machine learning, has led to notable interest in devising systems to automatically suggest commit messages.

\subsection{Commit Generation Approaches}
Commit message generation as a machine learning task is highly related to the NLP tasks of machine translation and summarization, although other approaches also exist. First, the meaning of the changes needs to be inferred, then the inferred information is translated into natural language and summarized to retain just a concise description. %

\subsubsection{Text-Based Approaches}
Many approaches treat commit message generation as a sequence-to-sequence task, for which NLP approaches based on deep neural networks can readily be adopted with just minimal changes. The diff is treated as an input text, and the commit message as the corresponding desired output text.
One of the first to apply deep learning to commit message generation, Jiang et al.~\cite{jiang_automatically_2017} used an Attentional RNN Encoder--Decoder, directly taken from a \textit{Neural Machine Translation} (NMT) framework, to translate the input diff to a commit message. 
Their dataset \textit{CommitGen\textsubscript{Data}} was created based on findings of their previous paper~\cite{jiang_towards_2017}, which studied the differences between auto-generated messages and human-written messages. They found that 82\% of human-written messages only contain one line of text and that the majority of these follow a ``verb+object'' format. 

Analyzing the results from Jiang et al.~\cite{jiang_automatically_2017}, Liu et al.~\cite{liu_neural-machine-translation-based_2018} found that the generated diffs on the test set share a high similarity to the ones seen during training. Additionally, they identified roughly 16\% as bot or trivial messages and subsequently removed them from the original dataset to generate a cleaned version. They proposed a nearest neighbor approach, \emph{NNGen\textsubscript{Model}}, which outperforms the NMT-based approach~\cite{jiang_automatically_2017} on CommitGen and on their NNGen dataset. They concluded that traditional retrieval approaches are also feasible for commit message generation, especially on a small dataset.

To improve the prediction of out-of-vocabulary words, Liu et al.~\cite{liu_generating_2019} proposed \emph{PtrGNCMsg}, an approach based on pointer-generation networks. They used the original CommitGen dataset~\cite{jiang_automatically_2017}, but also created an additional dataset on top of it, with another 1,000 Java projects not originally in CommitGen. Due to its small size, large overlap with CommitGen, and minimal use in later research, we do not discuss it further. 
Their results slightly outperform the CommitGen results.

Creating an LSTM-based encoder--decoder model, Pravilov et al.~\cite{pravilov_unsupervised_2021} made use of unsupervised pretraining tasks to learn meaningful embeddings for code edits. As one of their downstream tasks, they evaluated using the embeddings for commit message generation and were able to achieve a similar performance as the previous supervised models ~\cite{jiang_automatically_2017, liu_generating_2019}. 

After pretraining a Transformer model and learning contextualized code representations using a self-supervised code prediction task, Nie et al.~\cite{nie_coregen_2021} subsequently fine-tuned on the CommitGen dataset for commit message generation. %
Their results indicate that their suggested pretraining tasks are helpful for commit message generation, but they did not release any code to reproduce their results.

Elnaggar et al.\cite{elnaggar_codetrans_2021} transfer previous research~\cite{raffel_exploring_2020} into the area of NLP on source code. They 
used existing datasets made for different tasks, pre-training, transfer learning, and fine-tuning to develop their unified encoder--decoder Transformer model \textit{CodeTrans}. Their model was shown to outperform previous approaches across all six trained tasks, among them commit message generation.

Jung et al.~\cite{jung_commitbert_2021} presented \emph{CommitBERT}, a CodeBERT~\cite{feng_codebert_2020} model fine-tuned for commit message generation. The model uses a pretrained CodeBERT encoder and a freshly initialized Transformer-based decoder. 
To evaluate his approach, the author created a dataset based on the CodeSearchNet~\cite{husain_codesearchnet_2020} repository selection. The results showed that a pretrained Transformer-based model achieves higher scores compared to a non-pretrained one, but the model was not compared to other baseline models. %

\subsubsection{Structure-Based Approaches}
A number of recent approaches explore the use of \emph{Abstract Syntax Trees} (AST) instead of unstructured text input \cite{xu_commit_2019},\cite{liu_atom_2020}, \cite{dong_fira_2022}. A source code's structure can be parsed into a tree-based code representation often used by compilers. Analogous to natural language, the syntactic rules that define an AST differ amongst programming languages.

Xu et al.~\cite{xu_commit_2019} argued that code structure should explicitly be taken into consideration. %
They proposed \textit{CoDiSUM}, a custom encoder--decoder model with copying mechanisms. They used the CommitGen dataset, %
showing that the performance benefits from additionally modeling code structure. However, incorporating programming language-specific features requires more work to adapt the model to other languages than text-only approaches.

In a recent approach incorporating ASTs, Dong et al.~\cite{dong_fira_2022} focused on more fine-grained code change representations. They combine this with a Transformer decoder to generate the final commit messages. Although they provide an extensive evaluation, they train on the older and noisy CommitGen dataset. Additionally, their current implementation is limited to Java, due to their reliance on ASTs as inputs. Adapting to a new programming language would take significant effort, unlike with text-based approaches.

\section{Previous Datasets}
The choice of dataset greatly affects what aspects a model learns and hence how it performs across different benchmarks, even within the same task. In \autoref{table_datasets}, we give an overview of previous datasets~\cite{jiang_automatically_2017,liu_neural-machine-translation-based_2018,liu_generating_2019, loyola_neural_2017,xu_commit_2019, jung_commitbert_2021,tao_evaluation_2021} as well as concurrent work \cite{eliseeva_commit_2023, muennighoff_octopack_2023} with regard to a number of important attributes. We argue that all previous datasets on commit message generation disregard at least one -- often many more -- of the desiderata for a responsible, high-quality dataset. Out of these seven existing datasets, only three use any additional filters for quality control (CommitGen, NNGen, MCMD); only two are currently reproducible (MultiLang, MCMD); only one filters for permissively licensed repositories (CommitBERT). None have taken any privacy-preserving measures. In light of the ongoing academic~\cite{privacy}, legal~\cite{getty, copilotlitigation, rothchild}, public~\cite{guardiancopyright}, and political~\cite{aiact, italychatgpt} debates on privacy and license issues, these factors cannot be ignored. Another important aspect is that the growing need for reproducibility and extensive meta-data for training and evaluation datasets \cite{gebru_datasheets_2018,bender-friedman-2018-data}. Providing rich context on the data collection and processing steps can facilitate fairer, reproducible and privacy-aware datasets and can guide dataset creators to be more intentional in their design.

Another limitation is that most datasets are Java-only. Even the rare exceptions of datasets that do include multiple programming languages \cite{loyola_neural_2017,jung_commitbert_2021} generally do not verify the commit's programming language at the commit level, but merely at the level of entire repositories. It is therefore easy for commits in other programming languages to slip in, introducing noise. 
Finally, a limiting factor in previous datasets is the small repository selection. MCMD \cite{tao_evaluation_2021} draws its almost 2M commits from only 500 repositories, so that a large number of commits within training, validation, and test sets will not be independent of each other.
The authors find that the models' performance degrades by up to $73.41\%$ when they cleanly split projects between training and test set.

From the widespread lack of reproducibility and quality filtering all the way to license issues and small dataset sizes, it becomes obvious that all previous datasets for commit message generation fail to holistically satisfy all relevant desiderata for a modern evaluation dataset. 
This underscores the pressing need for a comprehensive, well-curated, and privacy-aware dataset that can serve as a benchmark for future research in commit message generation. The ideal dataset should not only be diverse in terms of programming languages but also be meticulously verified at the commit level to ensure the highest quality. Moreover, the dataset should be expansive enough to encompass a wide range of repositories, ensuring that the training and evaluation data are representative of real-world scenarios.

\begin{table*}[htbp]
    \caption{Comparison of existing datasets for commit message generation with regard to dataset best practices, quality, and programming languages. The last column shows our newly developed dataset CommitBench.}
    \label{table_datasets}
    \centering
    \resizebox{1\linewidth}{!}{
        \begin{tabular}{ c c c c c c c c c c c}
        \toprule
        \makecell{\textbf{Criteria/Dataset}}  & \makecell{\textbf{Commit}\\\textbf{Gen}} & \makecell{\textbf{NNGen}} & \makecell{\textbf{PtrGNC}\\\textbf{Msg}} & \makecell{\textbf{Multi}\\\textbf{Lang}} & \makecell{\textbf{CoDi}\\\textbf{Sum}} & \makecell{\textbf{Commit}\\\textbf{Bert}}  & \makecell{\textbf{MCMD}} & \makecell{\textbf{Commit}\\\textbf{Chronicle}} & \makecell{\textbf{Commit}\\\textbf{Pack}} & \makecell{\textbf{CommitBench}\\\textbf{(ours)}} \\
        \midrule
        Train   & 26,208 & 22,112 & 23,622 & 122,756 & 75,000 & 276,392 & 1,800,000 & 7,659,458 & 57,700,105 & 1,165,213 \\ 
        Valid   & 3,000  & 3,000  & 5,050  & 15,344  & 8,000  & 34,717  & 225,000 & 1,554,042 & \xmark & 249,689 \\ 
        Test    & 3,000  & 2,521  & 3,988  & 15,344  & 7,661  & 34,654  & 225,000 & 1,486,267 & \xmark & 249,688 \\ 
        Repositories  & 1,000 & 1,000 & 2,000 & 12 & 1,000  & 52k & 500 & 11.9k & 2.8m & 72k   \\
        Reproducible  & \xmark & \xmark & \xmark & \checkmark & \xmark & \xmark & \checkmark & \checkmark & \checkmark & \checkmark \\ 
        Deduplicated  & \xmark & \checkmark & \xmark & \checkmark & \xmark & \checkmark & \checkmark & \checkmark & \checkmark & \checkmark \\ 
        License-Aware Data Collection & \xmark & \xmark & \xmark & \xmark & \xmark & \checkmark & \xmark & \checkmark & \checkmark & \checkmark \\
        Published License  & \xmark  & MIT & \xmark  & \xmark  & \xmark  & Apache 2.0 & \xmark  & Various & MIT & CC BY-NC \\
        Additional Filtering & \checkmark & \checkmark & \xmark & \xmark &\xmark & \xmark & \checkmark & \checkmark & \checkmark & \checkmark\\
        Privacy Measures &\xmark  & \xmark & \xmark & \xmark & \xmark & \xmark  & \xmark & \checkmark & \xmark & \checkmark \\ 
        \begin{tabular}[c]{@{}l@{}}Programming\\ Languages\end{tabular} & Java & Java & Java & \begin{tabular}[c]{@{}r@{}}Java,\\ C++\\ Python\end{tabular} &Java & \begin{tabular}[c]{@{}r@{}}Java, Ruby\\ JavaScript, \\ Go, PHP, \\ Python\end{tabular} & \begin{tabular}[c]{@{}r@{}}Java, C\#, \\ C++, Python, \\ JavaScript\end{tabular} & \begin{tabular}[c]{@{}r@{}}20\\ Languages\end{tabular} & \begin{tabular}[c]{@{}r@{}}350\\ Languages\end{tabular} & \begin{tabular}[c]{@{}r@{}}Java, Ruby\\ JavaScript, \\ Go, PHP, \\ Python\end{tabular}\\
        \bottomrule
        \end{tabular}
}
\end{table*}

We proceed by reviewing the previous datasets used for commit message generation individually in further detail:

\paragraph{CommitGen\textsubscript{Data}~\cite{jiang_automatically_2017}} This is a small Java-only dataset that is based on the top 1,000 Java repositories. It uses a V-DO filter for qualitative commit filtering, but has several downsides such as trivial and bot-generated messages~\cite{etemadi_relevance_2020}, and duplicates. It also lacks reproducibility, and comes with fixed pretokenization, missing licenses, and a short sequence length for diffs and messages. Since this remains the most used dataset for commit message generation, we include it in our evaluation.

\paragraph{NNgen\textsubscript{Data}~\cite{liu_neural-machine-translation-based_2018}} A cleaned version\footnote{\url{ https://github.com/Tbabm/nngen.git}} of CommitGen\textsubscript{Data}. It removes trivial and bot commit messages, but otherwise suffers from the same limitations.

\paragraph{PtrGNCMsg\textsubscript{Data}~\cite{liu_generating_2019}} A dataset\footnote{\url{https://zenodo.org/record/2593787/files/ptrgn-commit-msg.zip}} that uses the same preprocessing as CommitGen\textsubscript{Data}, but includes the top 2,000 Java repositories, instead of the top 1,000. Since it is thus very similar to CommitGen\textsubscript{Data} in its shortcomings, we disregard it in our evaluation.

\paragraph{MutliLang~\cite{loyola_neural_2017}} A small dataset\footnote{\url{https://github.com/epochx/commitgen}} created from twelve hand-selected repositories. It includes multiple programming languages. Due to its extremely small size, and because it is not publicly available, we do not regard this dataset as a good basis for evaluation and hence exclude it in our own experiments.

\paragraph{CoDiSum\textsubscript{Data}~\cite{xu_commit_2019}} This is a Java-only dataset\footnote{\url{https://github.com/SoftWiser-group/CoDiSum/blob/master/data4CopynetV3.zip}}, which also uses the top 1,000 Java repositories as its data source. It is the only dataset that includes a file-level programming language filter, but lacks any other quality filtering, resulting in a larger dataset than datasets from similar data sources. Since its preprocessing is not customizable and the raw data is not provided, we disregard it in the evaluation.

\paragraph{CommitBert~\cite{jung_commitbert_2021}} The repository selection of CommitBert\footnote{\url{https://github.com/graykode/commit-autosuggestions/tree/master/experiment/dataset}} is based on that of \textit{CodeSearchNet}~\cite{husain_codesearchnet_2020}, which includes six programming languages.  It is pretokenized and only contains the actual code changes, not the context around them.

\paragraph{MCMD~\cite{tao_evaluation_2021}} A large dataset\footnote{\url{https://zenodo.org/record/5025758}} with five programming languages using the top 500 repositories on GitHub. Unlike other datasets, there is no sequence length filter, which makes the dataset unnecessary large. The pretokenization results in an average diff sequence length of 3,448. Processing sequences of that length constitutes a problem for most models. RNNs  struggle with vanishing gradients, while Transformer-based models are limited by the quadratic runtime of their attention mechanism, with most models unable to surpass 1,024 tokens (e.g., BERT can process a maximum of 512 Tokens).

While this research was underway, several studies also introduced new datasets for commit message generation, which we list here for completeness but do not further analyze or incorporate into the remainder of this paper:

\paragraph{CommitChronicle~\cite{eliseeva_commit_2023}}
A dataset\footnote{\url{https://zenodo.org/records/8189044}} with various filters, incorporating 10.7 million commits from almost 12,000 repositories across 20 programming languages. Uniquely, the authors focus on preserving commit history, arguing that it helps to generate commit messages.

\paragraph{CommitPack~\cite{muennighoff_octopack_2023}}
Two datasets that focus on leveraging commits for instruction fine-tuning of LLMs. Those datasets consist only of commits that change a single file. The datasets focus on instruction tuning, but could be used for commit message generation as well.

\textit{Merge Commits}, \textit{Reverse Commits}, and \textit{Binary Changes} are removed in all datasets.

\section{CommitBench}
\label{sec:commitbench}
\label{cha:dataset}

We present CommitBench, a new benchmark for commit message generation. Our focus is on providing researchers with a large, high-quality dataset that is compliant with ethical data standards such as considering licenses and privacy-sensitive content.

\subsection{Data Acquisition}
\label{sec:data_acquisition}
To compile a collection of commit messages, the first key step is to select source code repositories with relevant commits. For this, we draw on GitHub, by far the largest host of source code repositories.

With over 85 million new repositories on GitHub in 2022 alone\footnote{\url{https://octoverse.github.com/} (visited on June 1, 2023)},
it is not practical to incorporate every repository. Furthermore, there should be a first qualitative repository selection. This not only refines the dataset, but also ensures that the commit messages are representative of best practices in the industry. A consistently maintained repository from a large organization, which is potentially used by many others, likely has better commit messages than a small repository uploaded by an individual for mere exploratory use.

Our repository selection is based on that of \texttt{CodeSearchNet}~\cite{husain_codesearchnet_2020}. They collected all publicly available non-fork GitHub repositories that are used by at least one other project. 
Subsequently, all repositories without an MIT  license, which explicitly permits the re-distribution of parts of the repository as long as the original copy of the license is distributed, are removed.

This is important so that CommitBench can be publicly distributed. Due to the scale of the dataset, it would be unfeasible to ask every author for their explicit consent to consider their commits.

\subsection{Data Processing}
\label{sec:data_processing}
Within each downloaded repository, for each commit, we extract the commit message, diff, commit hash, project name, organization, commit author, and committer. The diff is obtained in textual form using the default \texttt{git diff} command, ensuring consistency in the extraction process. The commit hash is stored for reproducibility, so that together with the repository organization and name, the commit can be traced back to the original repository. Only commits with one parent are extracted, since having more than one indicates a merge commit, which does not add any source code of its own. To reduce the complexity of further data processing, we only store commits with a diff smaller than 1MB, as larger diffs typically do not change source code but data. Additionally, commits with excessively large diffs can skew the analysis and may not provide meaningful insights into typical coding practices or patterns.

\subsection{Filtering Methods}
\label{sec:filtering_methods}
Even among high-quality repositories, not every commit is well-suited as training or evaluation data. \autoref{table:filter_number} shows the amount of commits affected by the different filters. 
In the following, we describe filtering techniques invoked to ensure a high-quality dataset. It is worth noting that there is an inherent overlap between some filters. For instance, the sequence length filter for diffs and the source code language filter share an overlap of 7,809,633 commits. This is because many changes unrelated to programming code, such as configurations or documentation, can be quite verbose.

\begin{table}[htbp]
    \centering
    \caption{Commits affected by specific filters from the unfiltered dataset. Note that a single commit can be counted by multiple filters simultaneously.}
\label{table:filter_number}
        \begin{tabular}{l r}
        \toprule
        \textbf{Filter}  & \textbf{Number of Commits}  \\
        \midrule
        No filter & 23,284,371 \\
        Bot commits & 807,335 \\
        Binary commits & 4,026,669 \\
        Commit messages \textless ~8 & 8,070,122 \\
        Commit messages \textgreater ~128 & 783,232 \\
        Diffs \textgreater~512 & 13,296,820 \\
        Trivial messages & 270,330 \\
        Revert commits & 65,868 \\
        Commit message language & 193,695 \\
        Source code language & 11,252,158 \\
        Commits left after all filters & 1,664,590 \\
        \bottomrule
        \end{tabular}

\end{table}

\paragraph{Binary and File Mode Commits}
Commit sub-changes can be categorized into three types: binary changes, file mode changes, and textual changes.
A diff for a binary commit only indicates what file has been changed, but not what content changed. Due to the missing semantic information in the diff, it is impossible to discern enough meaning for a commit message from the commit diff, even for a professional developer, making commits for binary diffs unsuitable for conventional commit message generation. 

File mode changes, such as changes in a file's permission, represent a different challenge. While these changes are technically straightforward,
they often lack the contextual and semantic information required for meaningful commit message generation. For instance, a change in file permissions from ``read-only'' to ``read-write'' can easily be described in textual form using simple patterns, but the underlying reason might not be evident from the diff alone.

Hence, our work focuses on textual changes. Binary as well as file mode changes were excluded from the dataset.

\paragraph{Sequence Length}
As with most sequence-to-sequence tasks, the length of individual sequences is unevenly distributed, usually skewed towards shorter sequences. Current model architectures are limited with regard to the sequence length. They either have a predefined maximum number of tokens they can process, e.g., attention-based models like BERT~\cite{devlin_bert_2019} with 512 tokens, or their performance degrades with long inputs, e.g., for RNN-based models due to vanishing gradients. Our experiments showed that a code-based tokenizer, such as the one of CodeTrans\cite{elnaggar_codetrans_2021}, would reduce the sequence length for the diffs by 15\%, compared to the T5 tokenizer. Nevertheless, for a general benchmark focusing on existing models, we used the T5 tokenizer to determine the sequence length and removed diffs and messages longer than 512 tokens.
Additionally, commit messages shorter than eight tokens were removed, since such messages are usually of poor quality, lacking sufficient information on what was changed and why. We provide detailed statistics about the sequence length in \autoref{table:statistics}.

To evaluate long-sequence models, we additionally release a dataset version with a longer maximum sequence length of 2,048 tokens.

\begin{table}[htbp]
    \centering
    \caption{Statistical information on the collected commits before filtering for message and code diff length using the T5 tokenizer.}
    \label{table:statistics}
    \resizebox{1\linewidth}{!}{
    \begin{tabular}{l r r}
        \toprule
        \textbf{Statistic} & \textbf{Message Length} & \textbf{Code Diff Length} \\
        \midrule
        Mean & 27.52 & 4 129.75 \\
        Median & 11 & 662 \\
        Standard Deviation & 243.10 & 15 904.66 \\
        Variance & 59 098.21 & 252 958 201 \\
        Minimum Value & 0 & 0 \\
        Maximum Value & 619 941 & 499 044 \\
        Range & 619 941 & 499 044 \\
        25th Percentile & 6 & 286 \\
        50th Percentile & 11 & 662 \\
        75th Percentile & 20 & 2046 \\
        \bottomrule
    \end{tabular}
    }

\end{table}

\paragraph{Trivial Messages}
Trivial messages are commit messages that fail to convey significant meaning regarding what exactly was changed in the source code and why the change occurred. Training on these examples encourages non-informative commit messages. For CommitBench, those were derived by observing common patterns in non-informative commit messages over a range of projects. In the following, we list the regular expressions invoked to recognize trivial messages. 
\begin{itemize}
    \item \verb|update changelog v?[\d*\.]*|
    \item \verb|prepare version v?[\d*\.]*|
    \item \verb|bump version v?[\d*\.]*|
    \item \verb|modify makefile|
    \item \verb|update submodule .*|
\end{itemize}

\paragraph{Bot Commits}
Previous research~\cite{etemadi_relevance_2020} found that a non-negligible number of commits in the \texttt{CommitGen} dataset was created by development tools, which they called bot commits. 
Commits from bots were identified through the commit author's name, author's e-mail, committer's name, or committer's e-mail address. If any of these contained the word ``bot'', the commit was removed from the dataset. This filtering method, while straightforward, may not capture all bot-generated commits, but it constitutes a reasonable attempt at reducing their prevalence in the dataset.

\paragraph{Revert Commits}
A revert commit will undo a previous commit by creating a new commit that reverts all the previous commit changes.
Since Git automatically generates revert messages, they were excluded from the dataset. Revert commits can easily be identified by the usage of \textit{Revert Commit HASH} in the commit message. 

\paragraph{Commit Message Language}
Despite the focus on English projects,
there are still commits with non-English messages. %
We detected these with \texttt{fasttext}~\cite{joulin2016bag, joulin2016fasttext} language identification and removed them if its confidence for English is lower than 50\%. This ensures a more consistent dataset, minimizing potential biases or inaccuracies arising from non-English commit messages.

\paragraph{Source Code Language}
\label{sec:filter_source_code_language}
The original \texttt{CodeSearchNet} dataset filters out repositories that are not predominantly using one of the six considered programming languages. Despite this, many commits from this repository selection include changes in files that do not contain any source code written in those programming languages. %
Instead, many commits only change configuration or documentation files. As \autoref{table:filter_number} shows, this includes almost half the commits in the unfiltered data.  We therefore removed commits where less than 50\% of the associated file name extensions were among the following: \texttt{php}, \texttt{rb}, \texttt{go}, \texttt{js}, \texttt{py}, or \texttt{java}. This filtering criterion was established to maintain the dataset's focus on the primary programming languages and to ensure the relevance of the commits to code-centric tasks.

\paragraph{Duplicate Detection}
Duplicates, especially between the training and validation/test split, can have adverse effects on the evaluation of models~\cite{allamanis_adverse_2019}. CommitBench is deduplicated by removing samples with the same diff. Furthermore, the presence of duplicates can lead to overly optimistic performance metrics, as models might simply memorize patterns rather than genuinely learning to generalize. Hence, we hypothesize that the rigorous deduplication process in CommitBench is crucial for obtaining a realistic assessment of model capabilities.

\paragraph{Irrelevant Information}
Many commit messages include information that refers to a different platform, such as issue IDs and URLs. We replaced such information with unique tokens. Version numbers were also replaced in the commit message, since they can easily be extracted from the diff. This standardization ensures a more generalized representation, allowing models to focus on the core content of the commit message without being distracted by platform-specific details.

\paragraph{Privacy}
Extra effort was put into removing privacy-sensitive data, such as full names and e-mails. Full names were identified via named entity recognition using \textit{FLAIR}~\cite{akbik2019flair}, 
while e-mail addresses were identified via pattern matching. 
Both were replaced with special tags. We found that often the identified names and e-mails follow phrases such as \textit{Thanks to} or \textit{Contributed by}. We therefore removed all such lines in a commit message.
Additionally, we removed all author and committer names in the released dataset. %

\subsection{Final Dataset}
After filtering, the dataset is randomly partitioned into a 70\%/15\%/15\% split. We evaluate an alternative repository-based split strategy in \autoref{subsection:repo-split}.
Finally, the data is distributed as a \texttt{CSV} file, containing the project name, hash, commit message, diff, and split. This format has the advantage of retaining all pertinent data in a single file, facilitating data processing and sharing. For each commit and its commit message, we provide the respective programming language via file name extension analysis.

Our final dataset consists of 1,664,590 high-quality commits from 38,578 organizations and 71,676 unique projects, changing 1,781,414 files. To compare, applying our extensive filtering to MCMD, another large dataset, results in fewer than 400k total data samples, instead of its original 2.25M. Our commits stem from 169,316 unique authors and all repositories were downloaded on September 1, 2022. \autoref{table_languages} presents the programming language variety in the final dataset.

\begin{table}[ht!]
    \centering
         \caption{Overview of split by programming language in the diff for CommitBench. A single commit that contains changes in files with multiple programming languages will be counted towards each of them.}
    \label{table_languages}
     \begin{tabular}{c c c} 
        \toprule
     
     Language & \# Samples \\ 
     \midrule
     
     Java & 153,119  \\ 
     Ruby & 233,710  \\
     Go & 137,998 \\
     JavaScript & 373,598\\
     Python & 472,469 \\
     PHP & 294,394 \\
     \bottomrule
     \end{tabular}

    \end{table}

We have maintained transparency in our process by storing the used repositories in a list. To trace back to original commit data, one can use the hash, name, and repository. Given that repositories can be deleted or modified, we also hold a backup copy of all the repositories used, available on request for privacy reasons. These measures ensure CommitBench remains a dependable tool for research in the future.

\section{Experimental Setup}
\label{cha:experimental_setup}
The comparability of previous work has suffered from inconsistent metrics (e.g., BLEU coming in many variations). We remedy this by training and evaluating all models on a selection of previous commit message generation datasets as well as ours, simultaneously embedding CommitBench into the research landscape. In the following we describe the experimental setup used to answer our research questions:

\begin{itemize}
    \item[] \textbf{RQ1} How do various approaches, such as RNN-based, retrieval-based, and Transformer-based models, compare in terms of accuracy, diversity, and quality of generated commit messages?
    \item[] \textbf{RQ2} How does the split of a diverse dataset affect the accuracy of models?
    \item[] \textbf{RQ3} What are the effects of multi-language training compared to single-language training?
\end{itemize}

\subsection{Model Setup}
\label{sec:model_setup}
In our experiments, we benchmarked the following popular models, aiming at covering a diverse set of approaches: (1) CommitGen\textsubscript{Model} -- the first commit message generation model designed for this task, which is now a ubiquitous baseline. (2) NNGen\textsubscript{Model} -- the nearest-neighbor approach that came shortly after. (3) T5, a prominent Transformer model. (4) CodeTrans, a recent model with the same architecture as T5, but pretrained on source code tasks. For each Transformer-based model, we evaluated two different model sizes. \autoref{table:model_parameters} gives a comparison of the number of parameters of the evaluated models.

\begin{table}[h]
    \centering
    \caption{Comparison of the number of model parameters for the evaluated models.}
    \label{table:model_parameters}
        \begin{tabular}{lr}
            \toprule
             Models & \# Model Parameters\\
             
            \midrule
            \textbf{CommitGen} & 53 Million\\
            \textbf{NNGen} & 0 \\
            \textbf{T5\textsubscript{Small}} & 60 Million\\
            \textbf{T5\textsubscript{Base}} & 220 Million \\
            \textbf{CodeTrans\textsubscript{Small}} & 60 Million\\
            \textbf{CodeTrans\textsubscript{Base}} & 220 Million \\
            \bottomrule
        \end{tabular}

\end{table}

Our aim was to train the models in a similar setup to their original implementations. We therefore made use of the existing code repositories\footnote{NNGen: \url{https://github.com/Tbabm/nngen}\\ T5: \url{https://huggingface.co/t5-small} \\CodeTrans: \url{https://huggingface.co/SEBIS/code_trans_t5_small_transfer_learning_pretrain} \\ CommitGen's code repository has since been deleted.}. We took the models with default settings described in the respective papers.  For T5- and CodeTrans-based models, the hyperparameters given in \autoref{table:hyperparameters} were used. All experiments are reported based on a single run due to the extensive training times of large model sizes.

\begin{table}[ht]
    \centering
    \caption{Hyperparameters used for fine-tuning and evaluating T5- and CodeTrans-based models.}
    \label{table:hyperparameters}
        \begin{tabular}{lc}
            \toprule
             Parameter & Value\\
             
            \midrule
            Batch Size & 512 \\
            Optimization Method & AdamW \\
            Learning Rate & 5 $\times 10^{-5}$ with linear decay \\
            Gradient Clipping & 1.0 \\
            Dropout Rate & 0.1 \\
            Number of Epochs & 40 \\
            Beam Size & 5 \\
            \bottomrule
        \end{tabular}

\end{table}

While we provide an extensive overview of models used for commit generation, the list is non-exhaustive. This limitation is due to unavailable implementations or, in some cases, implementations that we estimated would need over two weeks to train.

\subsection{Datasets}
Apart from CommitBench, we included the original CommitGen dataset, due to its popularity in prior work. We further compared with NNGen\textsubscript{Data}, since it is a cleaned version of CommitGen\textsubscript{Data}. Lastly, we also incorporated MCMD, one of the largest commit message generation dataset.

\subsection{Evaluation Metrics}
Evaluating generated commit messages is a non-trivial task. Previous research~\cite{ jiang_automatically_2017,liu_generating_2019,pravilov_unsupervised_2021,elnaggar_codetrans_2021} has often used traditional machine translation metrics, such as BLEU~\cite{papineni_bleu_2001}, ROUGE-L~\cite{lin_rouge_2004}, and METEOR~\cite{denkowski_meteor_2014}, to compare the generated message to a reference message based on token overlap heuristics. This approach can lead to skewed results, as the set of valid commit messages for a single commit is often larger than in traditional translation tasks, yet only one reference message is provided.

Wang et al.~\cite{wang_quality_2021} highlight the challenges associated with ensuring the semantic relevance of generated commit messages. Their work underscores the need for a more refined evaluation mechanism that can discern between semantically relevant and irrelevant commit messages. To overcome this challenge we use 4 different evaluation metrics, briefly described in the following:

\textbf{BLEU} \cite{papineni_bleu_2001} is a common evaluation metric in machine translation. It compares a candidate output with a set of gold standard references. BLEU scores range from 0 to 1, with higher scores indicating better overlap and 1 indicating a perfect match with the reference. Due to the fact that numerous variations of BLEU exist, reported scores can vary widely. We chose the implementation by Post \cite{post_call_2018}, which is consistent with the WMT standard.

\textbf{ROUGE} \cite{lin_rouge_2004} was originally developed to evaluate summarization methods. It measures the overlap between the n-grams in the generated text and the reference text. We specifically consider the ROUGE-L variant, which focuses on the longest common subsequence, as it captures fluent and coherent sequence generation. ROUGE scores also range from 0 to 1, with higher scores indicating better overlap with the reference. We use a popular implementation\footnote{\url{https://pypi.org/project/rouge-score/}}.

\textbf{METEOR} \cite{banerjee-lavie-2005-meteor} is a machine translation metric originally designed to address problems with BLEU, such as its lack of synonym matching. It aims to obtain greater correlations with human evaluations. METEOR scores range from 0 to 1, with higher scores indicating better alignment with human judgments. We use the METEOR Universal variant \cite{denkowski_meteor_2014}.

\textbf{C-GOOD} \cite{tian_what_2022} uses a Bi-LSTM model trained on human-annotated commit messages to evaluate if a commit message contains information reflecting the \emph{why} and \emph{what} underlying the change. Conveying this in commit messages is crucial for developers to grasp the context and purpose of a commit. C-GOOD reports the percentage of good commit messages in the generated predictions in the range from 0 to 1, with higher percentages indicating better quality. Notably, it is the only metric that does not use references for evaluation. C-GOOD evaluates a commit message in isolation, without considering the actual code changes made in the commit. This means that the evaluation does not take into account how well the commit message aligns with the specific code modifications, potentially overlooking the relevance and accuracy of the message in the context of the changes made. We use the original script and data provided by the authors.\footnote{\url{https://github.com/WhatMakesAGoodCM/What-Makes-a-Good-Commit-Message}}

\section{Results}
\label{cha:results}
 \label{main_results}

\begin{table*}[htb]
    \centering
        \caption{Overview of test results comparing models each trained and tested on the same dataset, for four different datasets. Best evaluation results for each metric and dataset are in \textbf{bold}.}
    \label{table:1}
    \resizebox{1\linewidth}{!}{
        \begin{tabular}{lccccccc}
            \toprule
            \textbf{Dataset} & \textbf{Metric} & \textbf{CommitGen} & \textbf{NNGen} & \textbf{T5\textsubscript{Small}} & \textbf{T5\textsubscript{Base}} & \textbf{CodeTrans\textsubscript{Small}} & \textbf{CodeTrans\textsubscript{Base}} \\
            \midrule
            \multirow{4}{*}{\textbf{CommitGen}} 
            & METEOR & 0.418 & \textbf{0.428} & 0.357 & 0.397 & 0.385 & 0.404 \\
            & ROUGE-L & 0.373 & 0.382 & 0.354 & 0.409 & 0.405 & \textbf{0.433} \\
            & BLEU & 0.375 & \textbf{0.387} & 0.223 & 0.283 & 0.295 & 0.314 \\
            & C-GOOD & \textbf{0.418} & 0.417 & 0.381 & 0.344 & 0.354 & 0.361 \\
            \midrule
            \multirow{4}{*}{\textbf{NNGen}} 
            & METEOR & 0.309 & 0.325 & 0.291 & 0.326 & 0.315 & \textbf{0.339} \\
            & ROUGE-L & 0.246 & 0.270 & 0.243 & 0.300 & 0.297 & \textbf{0.330} \\
            & BLEU & 0.145 & 0.168 & 0.095 & 0.138 & 0.145 & \textbf{0.172} \\
            & C-GOOD & \textbf{0.439} & 0.438 & 0.401 & 0.374 & 0.375 & 0.383 \\
            \midrule
            \multirow{4}{*}{\textbf{MCMD}} 
            & METEOR & 0.199 & \textbf{0.289} & 0.229 & 0.243 & 0.236 & 0.250 \\
            & ROUGE-L & 0.130 & \textbf{0.202} & 0.139 & 0.166 & 0.163 & 0.181 \\
            & BLEU & 0.102 & \textbf{0.147} & 0.072 & 0.095 & 0.098 & 0.114 \\
            & C-GOOD & 0.489 & 0.538 & \textbf{0.578} & 0.555 & 0.556 & 0.570 \\
            \midrule
            \multirow{4}{*}{\textbf{CommitBench}} 
            & METEOR & 0.103 & 0.205 & 0.203 & 0.251 & 0.229 & \textbf{0.259} \\
            & ROUGE-L & 0.094 & 0.146 & 0.148 & 0.204 & 0.187 & \textbf{0.221} \\
            & BLEU & 0.023 & \textbf{0.096} & 0.021 & 0.047 & 0.033 & 0.053 \\
            & C-GOOD & 0.478 & 0.639 & \textbf{0.740} & 0.629 & 0.649 & 0.632 \\
            \bottomrule
        \end{tabular}
    }

\end{table*}

We provide a high-level overview of the overall performance of current commit message generation in \autoref{table:1} on the four commit message generation datasets. NNGen\textsubscript{Model} and the larger CodeTrans model obtain the strongest results across all datasets. The strong results of NNGen\textsubscript{Model} are remarkable, considering that it is a simple nearest-neighbor retrieval approach. An explanation could be that commits often follow similar use cases, enabling the model to retrieve many relevant neighbors. This phenomenon is likely most pronounced on the MCMD data, which is based on only 500 repositories, so that most commits in the test set match others from the same project observed during training. It is much less of an issue for NNGen\textsubscript{Data} and CommitBench, where for both NNGen\textsubscript{Model} performs noticeably worse than CodeTrans. This suggests that these results are due to more varied commits from distinct project sources. The results on CommitGen\textsubscript{Data} are the highest, but this likely stems from the models being able to rely on surface patterns.

It is worth noting that CodeTrans models are generally strong performers across the board, but do not consistently achieve the highest C-GOOD scores. This indicates that solely large-scale code-specific pre-training is not sufficient for commit message generation. Additionally, it is crucial to recognize that the measurement of ``good'' commit messages, as quantified by the C-GOOD metric, is not infallible. Like all metrics, C-GOOD has limitations, such as disregarding the actual code change, and might not capture all nuances of what constitutes a high-quality commit message in every context. Therefore, while it provides a valuable indication of quality, it should be interpreted in conjunction with other metrics and qualitative assessments to obtain a comprehensive understanding of a model's performance.

In summary, NNGen\textsubscript{Model} and CodeTrans show the strongest performance in commit message generation across four datasets. NNGen\textsubscript{Model} does especially well on the limited MCMD dataset due to its nearest-neighbor retrieval approach. However, CodeTrans outperforms NNGen\textsubscript{Model} on the other datasets, indicating a better handling of diverse commits.

\subsection{Output Diversity}
Since we frame commit message generation as a language generation task, we also evaluate how the choice of dataset affects the diversity of model outputs. More diverse outputs imply more flexible models that genuinely attempt to interpret the input and produce a custom output sequence reflecting it. If the outputs are not diverse while performance is high, this can be a reflection of easy-to-abuse patterns in the training data. We use Self-BLEU \cite{zhu_selfbleu_2018}, where a lower score means lower similarity with the other model outputs and thus a higher diversity.

\begin{figure}[htb]
    \centering
  \includegraphics[width=1\columnwidth]{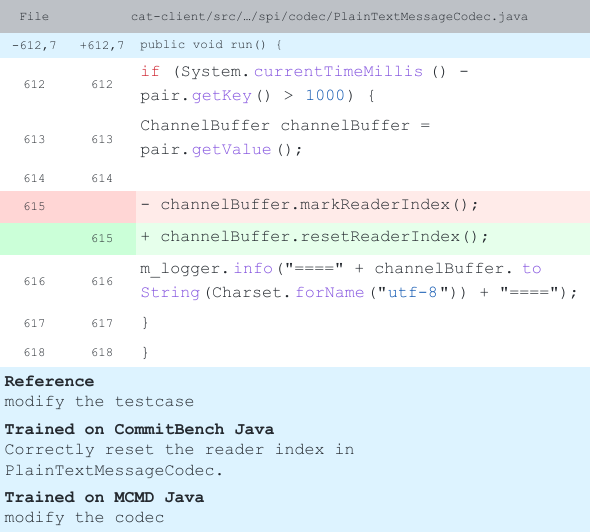}%
  \caption{Example of a diff, the reference commit message, and predicted commit messages from CodeTrans\textsubscript{Base} trained on the respective datasets. It can be observed that the better-generated commit message is further away from the reference message, which results in a lower evaluation score for this sample.}%
  \label{fig:example}
\end{figure}

\begin{table}
  \caption{Diversity of model outputs trained on all datasets, reported with Self-BLEU. Lower means more diverse. Best evaluation results in \textbf{bold}.}%
  \label{table:diversity}
    \center
  \resizebox{\linewidth}{!}{
  \begin{tabular}{ccccc}
            \toprule
             
Model  &CommitGen  & NNGen     & MCMD   & CommitBench   \\
\midrule
CommitGen       & 0.555     & 0.506     & 0.810  & 0.878\\
NNGen           & 0.490     & 0.412     & 0.345  & 0.348\\
T5\textsubscript{Small}        & 0.614     & 0.539     & 0.652  & 0.423\\
T5\textsubscript{Base}         & 0.579     & 0.510     & 0.591  & 0.388\\
CT\textsubscript{Small}        & 0.576     & 0.488     & 0.594  & 0.418\\
CT\textsubscript{Base}         & 0.539     & 0.468     & 0.569  & 0.382\\
            \midrule
Average         &0.559      &0.487      &0.594   &\textbf{0.473}\\
            \bottomrule
        \end{tabular}
}{%

}

\end{table}

The results are given in \autoref{table:diversity}.
We make two interesting observations. (1) Training on CommitBench results in models having the most diverse outputs. A close second is NNGen\textsubscript{Data}, tying into the observations made in the first experiment: that NNGen\textsubscript{Model} is more effective on CommitGen\textsubscript{Data} and MCMD, since it can rely on similar train and test sets. (2) CommitGen\textsubscript{Model}'s outputs on MCMD and CommitBench exhibit a low degree of diversity. We believe that, due to CommitGen\textsubscript{Model}'s limited complexity, the model is unable to capture the diversity of the larger CommitBench and MCMD datasets. 
The highly parametrized Transformer-based models can maintain diverse outputs on CommitBench, likely due to the large dataset size as well as the heterogeneous repository combination.

In short, sufficiently parametrized models trained on CommitBench have the most diverse output.

\subsection{Qualitative Analysis}
While manually reviewing the datasets, we noticed a significant discrepancy in the quality of commit messages. Specifically, in MCMD, many human-written commit messages do not adhere to the best practices of message composition. They often lack informativeness, being very short and sometimes vague. This is a serious problem, as models trained on MCMD tend to inherit these characteristics, often generating short and less informative commit messages. 

In contrast, since we employ a filtering mechanism that removes commits with fewer than eight tokens in CommitBench, the models trained on this dataset tend to produce longer and more detailed messages. This distinction becomes evident when we compare the outputs. We provide an illustrative example in \autoref{fig:example}. While the model trained on CommitBench gives a more informative and detailed output, the MCMD model's generation, albeit shorter, is closer to the human reference. This paradoxically would result in a higher evaluation score for the MCMD model. Such observations underscore the importance of dataset quality and curation. We take this as further validation of the need for extensive filtering and careful selection during dataset creation.

\subsection{Repository-based Split}
\label{subsection:repo-split}

The choice of evaluation split can significantly influence the performance of a model. One common approach to splitting the data is based on the repository from which a sample originates, instead of randomly splitting the samples on the commit-level. For instance, in the case of MCMD~\cite{tao_evaluation_2021}, the authors report that a change of splitting strategy from random to repository-based in their evaluation split led to a drastic 53\% drop in performance.

As illustrated in \autoref{table:split}, Transformer-based models were trained on CommitBench using two distinct evaluation splits. One version employed a random split, while the other adopted a repository-based division.

We observe that our dataset, CommitBench, is sufficiently diverse to mitigate the impact of splitting strategy. In comparison to MCMD, the difference between a random split and a repository-based split in CommitBench is negligible. This is underscored by the fact that, despite having a comparable number of samples, CommitBench sources its data from nearly 150 times more repositories than MCMD, with 72,000 repositories as opposed to MCMD's 500. Consequently, the potential for code overlap between the training and testing sets is substantially reduced in CommitBench, making it a more robust dataset.

\begin{table*}[htb]
    \centering
    \caption{Results of trained and evaluated T5-Based models on the CommitBench dataset. ``Random'' is the standard split, while ``Repo'' splits the dataset based on repositories.}
\label{table:split}
            \begin{tabular}{ccccccccc}
            \toprule
                     & \multicolumn{2}{c}{METEOR}& \multicolumn{2}{c}{ROUGE-L}& \multicolumn{2}{c}{BLEU}  &\multicolumn{2}{c}{C-GOOD}       \\
    \cmidrule(lr){2-3}
    \cmidrule(lr){4-5}
    \cmidrule(lr){6-7}
    \cmidrule(lr){8-9}
    \textbf{\textbf{Split}} & \textbf{\textbf{Random}} & \textbf{\textbf{Repo}} & \textbf{\textbf{Random}} & \textbf{\textbf{Repo}} & \textbf{\textbf{Random}} & \textbf{\textbf{Repo}} & \textbf{\textbf{Random}} & \textbf{\textbf{Repo}} \\
    \midrule
T5\textsubscript{Small}          & .203 & .186  & .148 & .123               & .021 & .014   & .740   & 0.814 \\
T5\textsubscript{Base}           & .251 & .220  & .204 & .176               & .047 & .029   & .629  &  0.669 \\
CodeTrans\textsubscript{Small}   & .229 & .0197 & .187 & .151               & .033 & .017   & .649   & 0.740 \\
CodeTrans\textsubscript{Base}    & .259 & .217  & .221 & .186               & .053 &.028    & .632  & 0.678  \\
    \bottomrule
    \end{tabular}

\end{table*}

The results in \autoref{table:split} reflect this and thus provide compelling insights into the robustness of CommitBench. While there is a noticeable performance drop in models evaluated using a repository-based split compared to a random split, the decline is not as pronounced as the one reported for MCMD. The slight performance drop in the repository-based split can be attributed to the inherent challenges of generating commit messages for previously unseen repositories. The fairly modest decrease in performance when the data is split based on repositories underscores the inherent diversity of CommitBench. 

In short, a more sophisticated splitting strategy for commit message generation is optional when the sourced data is sufficiently diverse.

\subsection{Multi-Language Impact}
Similar to natural languages, there exist a variety of programming languages. While some are more low-level, i.e., more hardware-related, others are higher-level languages with more abstract concepts. 
We hypothesize that, analogous to natural language models, the per-language performance of dedicated monolingual models in high-resource settings may be better than that of multi-language models. To verify this, we trained all T5-based models on both: each programming-language-specific subset of CommitBench as well as jointly on all samples.  %

\begin{table*}[htb]
        \centering
        \caption{Results when training the T5-based models on the six individual programming languages in CommitBench. Evaluation is then done on the language-specific subset of the test set and compared to the performance of the model trained on all languages on the same subsets. The results are reported as ROUGE-L scores. The better score within each mono- and multi-language comparison is colored blue. Best overall results for each language in \textbf{bold}.}
\label{table:mono}
        \resizebox{1\linewidth}{!}{
            \begin{tabular}{ccccccccccccc}
            \toprule
    Language                 & \multicolumn{2}{c}{Java}& \multicolumn{2}{c}{Python}& \multicolumn{2}{c}{Go}  &\multicolumn{2}{c}{JavaScript} &\multicolumn{2}{c}{PHP}  &\multicolumn{2}{c}{Ruby}      \\
    \cmidrule(lr){2-3}
    \cmidrule(lr){4-5}
    \cmidrule(lr){6-7}
    \cmidrule(lr){8-9}
    \cmidrule(lr){10-11}
    \cmidrule(lr){12-13}
    \textbf{\textbf{Model}} & \textbf{\textbf{One}} & \textbf{\textbf{All}} & \textbf{\textbf{One}} & \textbf{\textbf{All}} & \textbf{\textbf{One}} & \textbf{\textbf{All}} & \textbf{\textbf{One}} & \textbf{\textbf{All}} & \textbf{\textbf{One}} & \textbf{\textbf{All}} & \textbf{\textbf{One}} & \textbf{\textbf{All}}\\
    \midrule
    T5\textsubscript{Small}          & .130 & \cellcolor{blue!25}.140 & .151 & \cellcolor{blue!25}.157 & .131 & \cellcolor{blue!25}.150 & .116   & \cellcolor{blue!25}.126 & .124 & \cellcolor{blue!25}.139 & .170 & \cellcolor{blue!25}.186  \\
    T5\textsubscript{Base}           & .177 & \cellcolor{blue!25}.194 & .219 & .219 & .179 & \cellcolor{blue!25}.210 & .175  & \cellcolor{blue!25}.185 & .180 & \cellcolor{blue!25}.192 & .220 & \cellcolor{blue!25}.234 \\
    CodeTrans\textsubscript{Small}   & .167 & \cellcolor{blue!25}.180 & .189 & \cellcolor{blue!25}.203 & .157 & \cellcolor{blue!25}.180 & .147   & \cellcolor{blue!25}.165 & .163 & \cellcolor{blue!25}.177 & .209 & \cellcolor{blue!25}.222 \\
    CodeTrans\textsubscript{Base}    & .203 & \cellcolor{blue!25}\textbf{.214} & \textbf{.235} & \textbf{.235} & .209 & \cellcolor{blue!25}\textbf{.226} & .194  & \cellcolor{blue!25}\textbf{.201} & .209 & \cellcolor{blue!25}\textbf{.213} & .241 & \cellcolor{blue!25}\textbf{.249} \\
    \bottomrule
    \end{tabular}
    }

\end{table*}
The results are reported in \autoref{table:mono}. The multilingual setup consistently outperforms the monolingual one. We see several possible reasons. Learning from a diverse set of languages teaches the models fundamental principles all languages have in common.
Drawing from multiple programming languages appears to equip the models with a broader syntactic and semantic understanding, enabling them to handle diverse coding patterns more effectively. Another aspect is the larger number of training updates when combining all training examples. We find the latter to more likely be the key reason, supported by the fact that the differences in performance on Python are the smallest, which also has the most individual samples out of all languages (for an overview of samples per language, see \autoref{table_languages}). 

Overall, multilingual models consistently outperformed monolingual ones, likely due to their broader syntactic and semantic understanding from diverse language exposure and more extensive training updates.

\section{Conclusion}
This paper presents CommitBench, a modern and comprehensive dataset for commit message generation. Reviewing existing datasets for the task, we identify critical weaknesses with regard to the quality of the data. Based on a discussion of best practices for dataset creation, we construct a number of preprocessing and filtering techniques to address these shortcomings and compile CommitBench. We unify previous research by comparing reproducible approaches for commit message generation, providing a consistent set of metrics across all datasets and models. Simultaneously, we embed CommitBench into the current research landscape by evaluating it against previous datasets, showing that training on CommitBench leads to models producing more diverse outputs and generating new baseline values for reference in future work. We show that fine-tuning Transformer-based models outperforms other approaches, and cross-programming-language training yields improved results.

A limitation of automatic commit message generation is the need for a metric to measure commit message quality. As we have shown, existing commit messages are occasionally not a good evaluation reference. In addition, while the correct answers for translation tasks are often straightforward, there can be fairly diverse yet valid commit messages. Furthermore, the context in which a commit is made, including the developer's intent and the broader project goals, can greatly influence the ideal message. This adds another layer of complexity to the evaluation, as understanding this context is often beyond the scope of automated systems. However, a direct human evaluation would require a large pool of programming experts on diverse software topics who would need to spend substantial time studying the broader context of each commit in a repository in order to adequately judge the quality of commit messages.

To overcome the previous limitations, future work could incorporate the evaluation into programmers' workflows with an integrated feedback loop. This could aid in evaluating datasets and further enhance trainable metrics like C-Good. Further, CommitBench could be extended to a much larger set of programming languages and repositories. Additionally, evaluating AST-based commit message generation approaches would be viable, though they require significant work since those approaches are usually programming-language-specific and would need to be adapted for every single evaluated language. Another promising research direction is using large language models, which we briefly tried but got inconsistent results, leaving space for further exploration. Additional contextual information, such as issues, could further help understand a developer's intent.

In conclusion, we hope that our public release of CommitBench at \url{https://github.com/maxscha/commitbench}
will enable new research on these and other aspects of commit message generation.

\bibliographystyle{IEEEtran}
\bibliography{IEEEabrv,custom}

\newpage

\appendices

\section{Data Statement for CommitBench}
\label{app:dataset}
\subsection{HEADER}
\begin{itemize}
    \item Dataset Title: CommitBench
    \item Dataset Curator: Maximilian Schall, Hasso Plattner Institute
    \item Dataset Version: 1.0, 01.01.2024
    \item Data Statement Author: Tamara Czinczoll, Hasso Plattner Institute
    \item Data Statement Version: 1.0, 16.01.2023
    \item Data URL: \url{https://zenodo.org/records/10497442},\url{https://huggingface.co/datasets/maxscha/commitbench} 
    \item Code URL: \url{https://github.com/maxscha/commitbench}
\end{itemize}

\subsection{EXECUTIVE SUMMARY}
We provide CommitBench as an open-source, reproducible and privacy- and license-aware benchmark for commit message generation. The dataset is gathered from github repositories with licenses that permit redistribution. We provide six programming languages, Java, Python, Go, JavaScript, PHP and Ruby. The commit messages in natural language are restricted to English, as it is the working language in many software development projects. The dataset has 1,664,590 examples that were generated by using extensive quality-focused filtering techniques (e.g. excluding bot commits). Additionally, we provide a version with longer sequences for benchmarking models with more extended sequence input.

\subsection{CURATION RATIONALE}
We created this dataset due to quality and legal issues with previous commit message generation datasets. Given a git diff displaying code changes between two file versions, the task is to predict the accompanying commit message describing these changes in natural language. We base our GitHub repository selection on that of a previous dataset, CodeSearchNet, but apply a large number of filtering techniques to improve the data quality and eliminate noise. Due to the original repository selection, we are also restricted to the aforementioned programming languages. It was important to us, however, to provide some number of programming languages to accommodate any changes in the task due to the degree of hardware-relatedness of a language. The dataset is provides as a large CSV file containing all samples. We provide the following fields: Diff, Commit Message, Hash, Project, Split.

\subsection{DOCUMENTATION FOR SOURCE DATASETS}
\label{sec:source_data}
Repository selection based on CodeSearchNet, which can be found under \url{https://github.com/github/CodeSearchNet}.

\subsection{LANGUAGE VARIETIES}
Since GitHub hosts software projects from all over the world, there is no single uniform variety of English used across all commit messages. This means that phrasing can be regional or subject to influences from the programmer's native language. It also means that different spelling conventions may co-exist and that different terms may used for the same concept. Any model trained on this data should take these factors into account. For the number of samples for different programming languages, see Table~\ref{table_languages}.
\begin{table}[ht!]
     \caption{Overview of split by programming language for CommitBench. A single commit which contains multiple programming language will be counted towards each of them.}
    \label{table_languages_2}
    \centering
     \begin{tabular}{c c c} 
        \toprule
     
     Language & \# Samples \\ 
     \midrule
     
     Java & 153,119  \\ 
     Ruby & 233,710  \\
     Go & 137,998 \\
     JavaScript & 373,598\\
     Python & 472,469 \\
     PHP & 294,394 \\
     \bottomrule
     \end{tabular}
    \end{table}

\subsection{SPEAKER DEMOGRAPHIC}
Due to the extremely diverse (geographically, but also socio-economically) backgrounds of the software development community, there is no single demographic the data comes from. Of course, this does not entail that there are no biases when it comes to the data origin. Globally, the average software developer tends to be male and has obtained higher education. Due to the anonymous nature of GitHub profiles, gender distribution information cannot be extracted.

\subsection{ANNOTATOR DEMOGRAPHIC}
Due to the automated generation of the dataset, no annotators were used.

\subsection{SPEECH SITUATION AND CHARACTERISTICS}
The public nature and often business-related creation of the data by the original GitHub users fosters a more neutral, information-focused and formal language. As it is not uncommon for developers to find the writing of commit messages tedious, there can also be commit messages representing the frustration or boredom of the commit author. While our filtering is supposed to catch these types of messages, there can be some instances still in the dataset.

\subsection{PREPROCESSING AND DATA FORMATTING}
See Section~\ref{cha:dataset} for all preprocessing steps. We do not provide the un-processed raw data due to privacy concerns, but it can be obtained via CodeSearchNet (see Section~\ref{sec:source_data}) or requested from the authors.

\subsection{CAPTURE QUALITY}
While our dataset is completely reproducible at the time of writing, there are external dependencies that could restrict this. If GitHub shuts down and someone with a software project in the dataset deletes their repository, there can be instances that are non-reproducible. 

\subsection{LIMITATIONS}
While our filters are meant to ensure a high quality for each data sample in the dataset, we cannot ensure that only low-quality examples were removed. Similarly, we cannot guarantee that our extensive filtering methods catch all low-quality examples. Some might remain in the dataset. Another limitation of our dataset is the low number of programming languages (there are many more) as well as our focus on English commit messages. There might be some people that only write commit messages in their respective languages, e.g., because the organization they work at has established this or because they do not speak English (confidently enough). Perhaps some languages' syntax better aligns with that of programming languages. These effects cannot be investigated with CommitBench. 

Although we anonymize the data as far as possible, the required information for reproducibility, including the organization, project name, and project hash, makes it possible to refer back to the original authoring user account, 
since this information is freely available in the original repository on GitHub. 

\subsection{METADATA}
\begin{itemize}
    \item License:  Dataset under the CC BY-NC 4.0\footnote{\url{https://creativecommons.org/licenses/by-nc/4.0/}} license, code under the MIT license 
\end{itemize}

\subsection{DISCLOSURES AND ETHICAL REVIEW}
While we put substantial effort into removing privacy-sensitive information, our solutions cannot find 100\% of such cases. This means that researchers and anyone using the data need to incorporate their own safeguards to effectively reduce the amount of personal information that can be exposed.

\subsection{OTHER}

\subsection{ABOUT THIS DOCUMENT}
A data statement is a characterization of a dataset that provides context to allow developers and users to better understand how experimental results might generalize, how software might be appropriately deployed, and what biases might be reflected in systems built on the software.

This data statement was written based on the template for the Data Statements Version 2 schema. The template was prepared by Angelina McMillan-Major, Emily M. Bender, and Batya Friedman and can be found at https://techpolicylab.uw.edu/data-statements/ and was updated from the community Version 1 Markdown template by Leon Dercyznski.

\section{Data Examples}\label{app:examples}
\begin{figure}[ht]
    \includegraphics[width=0.9\columnwidth]{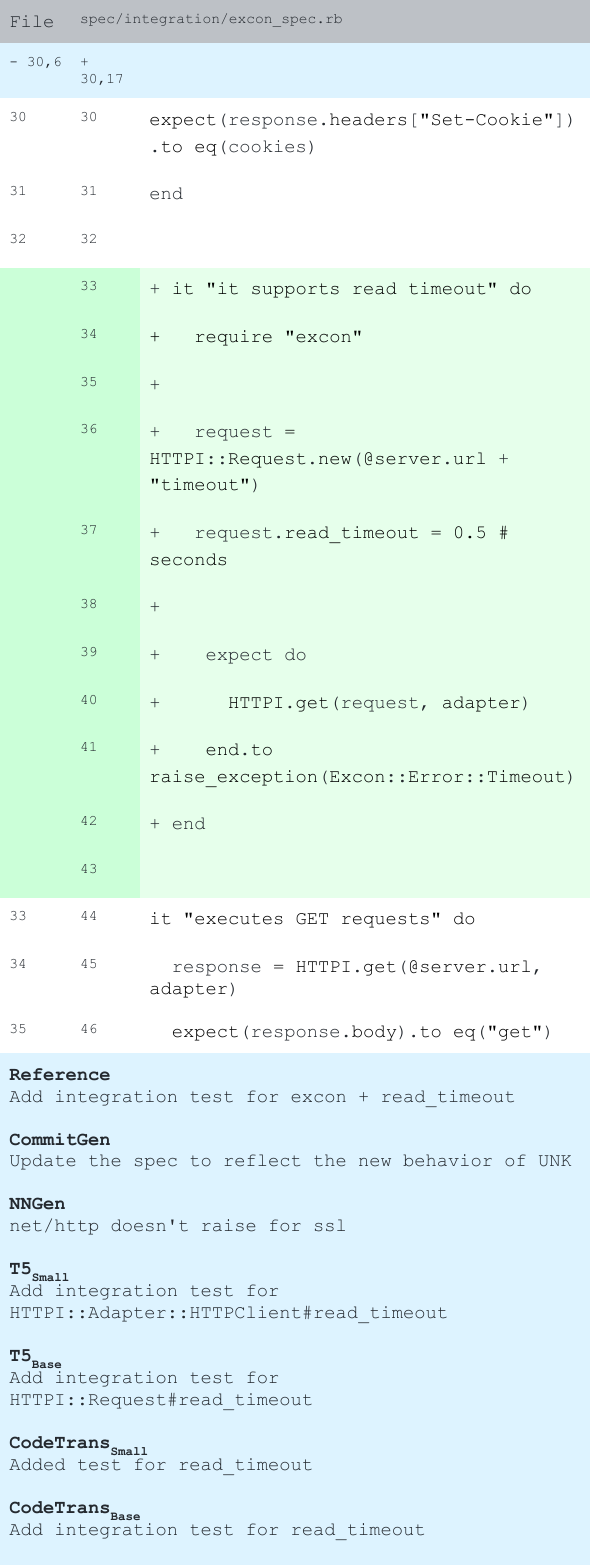}
    
    \caption{Sample of CommitBench-Test data, with predictions from CommitGen, NNGen, T5\textsubscript{Small}, T5\textsubscript{Base}, CodeTrans\textsubscript{Small}, and CodeTrans\textsubscript{Base}}
    \label{fig:sample_1}
\end{figure}

\begin{figure}[ht]
    \includegraphics[width=0.75\columnwidth]{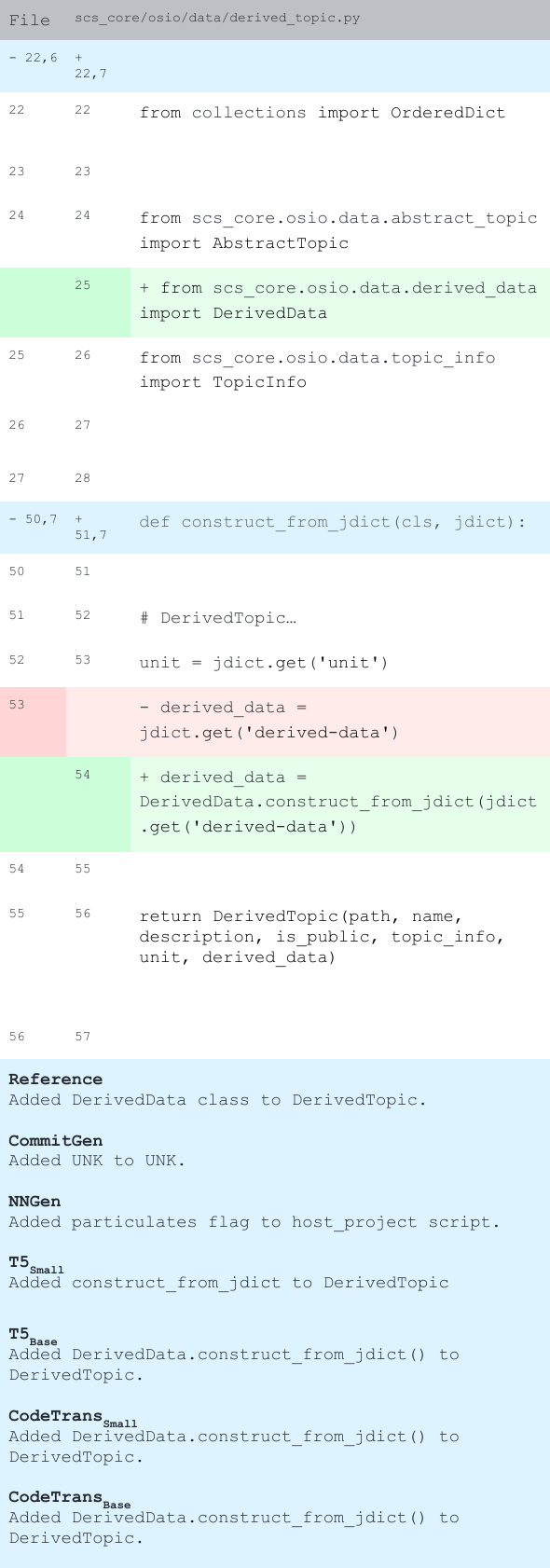}
    
    \caption{Sample of CommitBench-Test data, with predictions from CommitGen, NNGen, T5\textsubscript{Small}, T5\textsubscript{Base}, CodeTrans\textsubscript{Small}, and CodeTrans\textsubscript{Base}}
    \label{fig:sample_2}
\end{figure}

\begin{figure}[ht]
    \includegraphics[width=1\columnwidth]{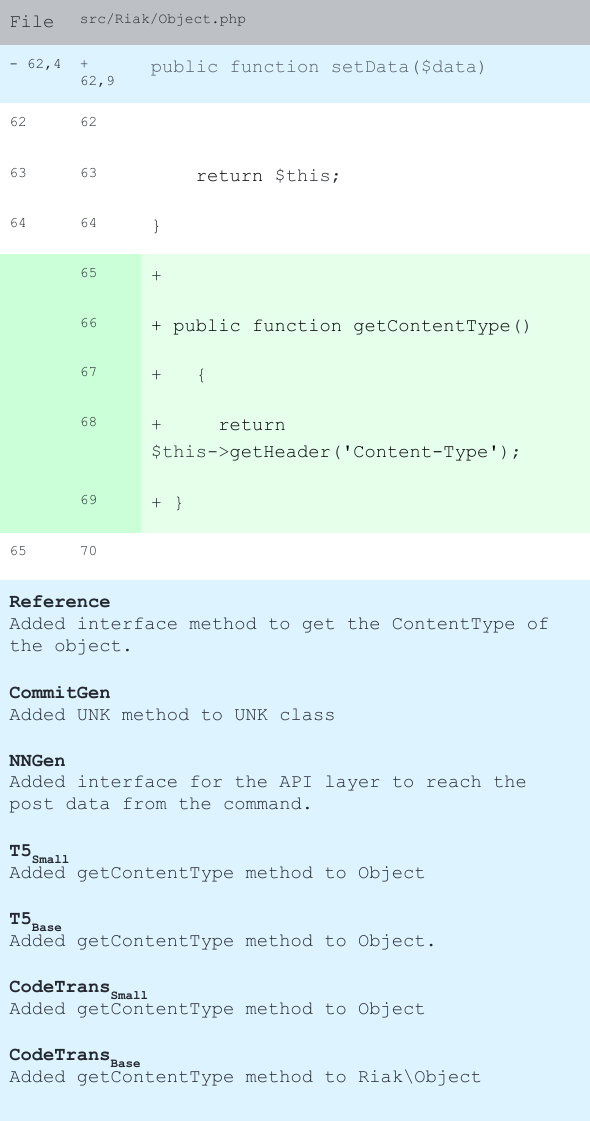}
    
    \caption{Sample of CommitBench-Test data, with predictions from CommitGen, NNGen, T5\textsubscript{Small}, T5\textsubscript{Base}, CodeTrans\textsubscript{Small}, and CodeTrans\textsubscript{Base}}
    \label{fig:sample_3}
\end{figure}

We show a selection of data examples of the CommitBench dataset in Figures \ref{fig:sample_1}, \ref{fig:sample_2}, and \ref{fig:sample_3}.

\section{Computational Budget}
As with most sequence-to-sequence tasks, the length of individual sequences is unevenly distributed, usually skewed towards shorter sequences. Current model architectures are limited with regard to the sequence length. Either they have a predefined maximum number of tokens they can process, e.g., attention-based models like BERT~\cite{devlin_bert_2019} with 512 tokens, %
or their performance degrades with long inputs, e.g., for RNN-based models due to vanishing gradients. To account for model and computational limitations we used the T5 tokenizer in our experiments to determine the sequence length, and removed diffs and messages longer than 512 tokens.

For CommitGen\textsubscript{Model} and NNGen\textsubscript{Model}, we used the hyperparameters reported by their original papers. CommitGen was trained on one NVIDIA A6000 with 50GB VRAM for 38h on CommitBench, while NNGen does not require a GPU and took 3h to compute the predictions for CommitBench.

These models were fine-tuned on four NVIDIA A100 with 40GB VRAM each. The \textit{Small} models took 12h for fine-tuning on CommitBench, while the \textit{Base} models took 35h for fine-tuning on CommitBench.

\end{document}